\documentclass{article}
\usepackage{spconf,amsmath,graphicx}
\include{pythonlisting}
\usepackage{algorithm}
\usepackage[noend]{algpseudocode}
\usepackage{slashbox}
\usepackage{multirow}
\usepackage{url}
\usepackage{epstopdf}

\makeatletter

\makeatletter
\newcommand{\HEADER}[1]{\ALC@it\underline{\textsc{#1}}\begin{ALC@g}}
\newcommand{\ENDHEADER}{\end{ALC@g}}
\makeatother

\title{Batch-normalized joint training for DNN-based \\ distant speech recognition}
\name{Mirco Ravanelli$^{\S,\star}$\thanks{$^\star$This work was done while the author was visiting the Montreal Institute for Learning Algorithms (MILA)  and was supported by the FBK mobility programme.}, Philemon Brakel$^{\dagger}$, Maurizio Omologo$^{\S}$, Yoshua Bengio$^{\dagger}$}

\address{$^{\S}$Fondazione Bruno Kessler, Trento, Italy \\ 
$^{\dagger}$Universit\'e de Montr\'eal, Montr\'eal, Canada} 
\begin{document}
%
\maketitle
\begin{abstract}
Improving distant speech recognition is a crucial step towards flexible human-machine interfaces. Current technology, however, still exhibits a lack of robustness, especially when adverse acoustic conditions are met.  Despite the significant progress made in the last years on both speech enhancement and speech recognition, one potential limitation of state-of-the-art technology lies in composing modules that are not well matched because they are not trained jointly. 

To address this concern,  a promising approach consists in concatenating a speech enhancement and a speech recognition deep neural network and  to jointly update their parameters as if they were within a single bigger network.
Unfortunately, joint training can be difficult because the output distribution of the speech enhancement system may change substantially during the optimization procedure. The speech recognition module would have to deal with an input distribution that is non-stationary and unnormalized. To mitigate this issue, we propose a joint training approach based on a fully batch-normalized architecture.

Experiments, conducted using different datasets, tasks and acoustic conditions, revealed that the proposed framework significantly overtakes other competitive solutions, especially in challenging environments.
\end{abstract}
\section{Introduction}
\label{sec:intro}
Automatic Speech Recognition (ASR) \cite{lideng}, thanks to the substantial performance improvement achieved with modern deep learning technologies \cite{Goodfellow-et-al-2016-Book}, has recently been applied in several fields, 
and it is currently  used by millions of users worldwide. 
Nevertheless, most state-of-the-art systems are still based on close-talking solutions, forcing the user to speak very close to a microphone-equipped device. 
It is easy to predict, however, that in the future users will prefer to relax the constraint of handling or wearing any device to access speech recognition services, requiring technologies able to cope with a distant-talking (far-field) interaction. 

In the last decade, several efforts have been devoted to improving Distant Speech Recognition (DSR) systems. Valuable examples include the AMI/AMIDA projects \cite{ami}, who were focused on automatic meeting transcription, DICIT \cite{dicit_1} which investigated voice-enabled TVs and, more recently, DIRHA which addressed speech-based domestic control. 
The progress in the field was also fostered by the considerable success of some international challenges such as CHiME \cite{chime,chime3} and REVERB \cite{revch_full}.

Despite the great progress made in the past years, current systems still exhibit a significant lack of robustness to acoustic conditions characterized by non-stationary noises and acoustic reverberation \cite{adverse}. 
To counteract such adversities, even the most recent DSR systems \cite{nakatani} 
must rely on a combination of several interconnected technologies, including for instance speech enhancement \cite{BrandWard}, speech separation \cite{bss}, acoustic event detection and classification \cite{aed1,eusipco}, speaker identification \cite{Beigi}, speaker localization \cite{gcf,hscma}, just to name a few.

A potential limitation of most current solutions lies in the weak matching and communication between the various modules being combined.
For example, speech enhancement and speech recognition are often designed independently and, in several cases, the enhancement system is tuned according to metrics which are not directly correlated with the final ASR performance. 

An early attempt to mitigate this issue was published in \cite{limabeam}. In LIMABEAM, the goal was to tune the parameters of a microphone array beamformer by maximizing  the likelihood  
obtained through a GMM-based speech recognizer. Another approach was proposed in \cite{droppo}, where a front-end for feature extraction and a GMM-HMM back-end were jointly trained using maximum mutual information. 

An effective integration between the various systems, however, was very difficult for many years, mainly due to the different nature of the technologies involved at the various steps.  
Nevertheless, the recent success of deep learning has not only largely contributed to the substantial improvement of the speech recognition part of a DSR system \cite{pawel2,hain,dnn_rev,dnn_rev2,dnn3,rav_in14,ravanelli15}, but has also enabled the development of competitive DNN-based speech enhancement solutions \cite{dnn_se1,dnn_se2,dnn_se3}. 
Within the DNN framework, one way to achieve a fruitful integration of the various components is joint training.  
The core idea is to pipeline a speech enhancement and a speech recognition deep neural networks and to jointly update their parameters as if they were within a single bigger network. Although joint training for speech recognition is still an under-explored research direction, such a paradigm is progressively gaining more attention and some interesting works in the field have been recently published \cite{joint2,joint1,joint3,joint6,joint7,joint4,joint5}. 

In this paper, we contribute to this line of research by proposing an approach based on joint training of a speech enhancement and a speech recognition DNN coupled with batch normalization in order to help making one network less sensitive to changes in the other. Batch normalization \cite{batchnorm}, which has recently been proposed in the machine learning community, has been shown crucial to significantly improve both the convergence and the performance of the proposed joint training algorithm.
Differently to previous works \cite{joint1,joint3}, thanks to batch normalization, we are able to effectively train the joint architecture even without any pre-training steps. 
Another interesting aspect concerns a deeper study of a gradient weighting strategy, which ended up being particularly effective to improve performance.

The experimental validation has been carried out in a distant-talking scenario considering different training datasets, tasks and acoustic conditions.

\section{Batch-normalized joint training}

\begin{figure}[t!]
\centering
\includegraphics[width=0.42\textwidth]{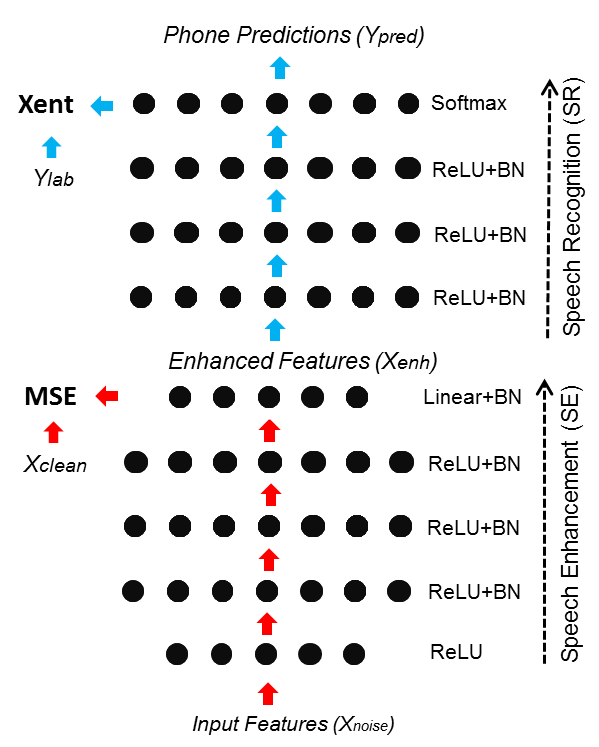}
\caption{The DNN architecture proposed for joint training.}
\label{fig:arch}
\end{figure}

The proposed architecture is depicted in Fig.~\ref{fig:arch}. A bigger joint DNN is built by concatenating a speech enhancement and a speech recognition MLP. The speech enhancement DNN is fed with the noisy features $x_{noise}$ gathered within a context window and tries to reconstruct at the output the original clean speech (regression task). 
The speech recognition DNN is fed by the enhanced features $x_{enh}$ estimated at the previous layer and performs phone predictions $y_{pred}$ at each frame (classification task). The architecture of Fig. \ref{fig:arch} is trained with the algorithm described in Alg. \ref{alg}.

\label{sec:format}
The basic idea is to perform a forward pass, compute the loss functions at the output of each DNN (mean-squared error for speech enhancement and negative multinomial  log-likelihood for speech recognition), compute and weight the corresponding gradients, and back-propagate them.
In the joint training framework, the speech recognition gradient is also back-propagated through the speech enhancement DNN. Therefore, at the speech enhancement level, the parameter updates not only depend on the speech enhancement cost function but also on the speech recognition loss, as shown by Eq.~\ref{eq:updates}:
 \begin{equation}
 \theta_{SE} \gets \theta_{SE}- lr * (g_{SE}+\lambda g_{SR}) \,.
 \label{eq:updates}
 \end{equation}
In Eq.~\ref{eq:updates}, $\theta_{SE}$ are the parameters of the speech enhancement DNN, $g_{SE}$ is the gradient of such parameters computed from the speech enhancement cost function (mean squared error), while $g_{SR}$ is the gradient of $\theta_{SE}$ computed from the speech recognition cost function (multinomial log-likelihood). Finally, $\lambda$ is a hyperparameter for weighting $g_{SR}$ and $lr$ is the learning rate.

The key intuition behind joint training is that since the enhancement process is in part guided by the speech recognition cost function, the front-end would hopefully be  able to provide enhanced speech which is more suitable and discriminative for the subsequent speech recognition task. 

From a machine learning perspective, this solution can also be considered as a way of injecting a useful task-specific prior knowledge into a deep neural network.
On the other hand, it is well known that training deep architectures is easier when some hints are given about the targeted function \cite{know_matter}. 
As shown previously \cite{know_matter}, such prior knowledge becomes progressively more precious as the complexity of the problem increases and can thus be very helpful for a distant speech recognition task. Similarly to the current work, in \cite{know_matter,Romero-et-al-ICLR2015-small} a task-specific prior knowledge has been injected into an intermediate layer of a DNN for better addressing an image classification problem.

In our case, we exploit the prior assumption that to solve our specific problem, it is reasonable to first enhance the features and, only after that, perform the phone classification.
Note that this is certainly not the only way of solving the problem, but among all the possible functions able to fit the training data, we force the system to choose from a more restricted subset, potentially making training easier. 
On the other hand, good prior knowledge is helpful to defeat the curse of dimensionality, and 
a complementary view is thus to consider the proposed joint training as a regularizer. 
According to this vision, the weighting parameter $\lambda$ of Eq. \ref{eq:updates} can be regarded as a regularization hyperparameter, as will be better discussed in Sec. \ref{sec:gw}. 
\begin{algorithm}[t!]
\caption{Pseudo-code for joint training}
\label{alg}
\begin{algorithmic}[1]
\State \textbf{DNN initialization} 
\For {i in minibatches}
 \State \textbf{Forward Pass:} 
 \State Starting from the input layer do a forward pass
 \State (with batch normalization) through the networks.
  \State \textbf{Compute SE Cost Function:} 
  \State $MSE_i=\frac{1}{N}\sum_{n=1}^{N}(x_{enh}^i-x_{clean}^i)^2$
  \State \textbf{Compute SR Cost Function:}
  \State $NLL_i=-\frac{1}{N}\sum_{n=1}^{N}y_{lab}^i log(y_{pred}^i)$ 
  \State \textbf{Backward Pass:}
  \State Compute the grad. $g_{SE}^i$ of $MSE_i$ and backprogate it.
  \State Compute the grad. $g_{SR}^i$ of $NLL_i$ and backprogate it.
  \State \textbf{Parameters Updates:}
   \State  $\theta_{SE}^i \gets \theta_{SE}^i - lr * (g_{SE}^i+\lambda g_{SR}^i)$
    \State  $\theta_{SR}^i \gets \theta_{SR}^i - lr * g_{SR}^i$
\EndFor
\State Compute NLL on the development dataset
\If {$NLL_{dev} < NLL_{dev}^{prev}$}
  \State Train for another epoch (go to 2) 
\Else
 \State Stop Training
\EndIf
\end{algorithmic}
\end{algorithm}
\subsection{Batch normalization} \label{sec:batchnorm}
Training DNNs is complicated by the fact that the distribution of each layer's inputs changes during training, as the parameters of the previous layers change.
This problem, known as \textit{internal covariate shift}, slows down the training of deep neural networks. 
Batch normalization \cite{batchnorm}, which has been recently proposed in the machine learning community, addresses this issue by  normalizing the mean and the variance of each layer for each training mini-batch, and back-propagating through the normalization step. It has been long known 
that the network training converges faster if its inputs are properly normalized \cite{yann} and, in such a way, batch normalization extends the normalization to all the layers of the architecture. However, since a per-layer normalization may impair the model capacity, a trainable scaling parameter $\gamma$ and a trainable shifting parameter $\beta$  are introduced in each layer to restore the representational power of the network. 

The idea of using batch normalization for the joint training setup is motivated by a better management of the internal covariate shift 
problem, which might be crucial when training our (very) deep joint architecture.
As will be shown in Sec.\ \ref{sec:bn_exp}, batch normalization allows us to significantly improve the performance of the system, to speed-up the training, and to avoid any time-consuming pre-training steps.

Particular attention should anyway be devoted to the initialization of the $\gamma$ parameter. Contrary to \cite{batchnorm}, where it was initialized to unit variance ($\gamma=1$), in this work we have observed better performance and convergence properties with a smaller variance initialization ($\gamma=0.1$).
A similar outcome has been found in \cite{initbn}, where fewer vanishing gradient problems are empirically observed with small values of $\gamma$ in the case of recurrent neural networks.

\subsection{System details}
The features considered in this work are standard 39 Mel-Cepstral Coefficients (MFCCs) computed every 10 ms with a frame length of 25 ms. The speech enhancement DNN is fed with a context of 21 consecutive frames and predicts (every 10 ms) 11 consecutive frames of enhanced MFCC features. The idea of predicting multiple enhanced frames was also explored in \cite{joint3}. 
All the layers used Rectified Linear Units (ReLU), except for the output of the speech enhancement (linear) and the output of speech recognition (softmax).
Batch normalization \cite{batchnorm} is employed for all the hidden layers, while dropout \cite{dropout} is adopted in all part of the architecture, except for the output layers. 

The datasets used for joint training are obtained through a contamination of clean corpora (i.e., TIMIT and WSJ) with noise and reverberation. 
The labels for the speech enhancement DNN (denoted as $x_{clean}$ in Alg.1) are the MFCC features of the original clean datasets.
The labels for the speech recognition DNN (denoted as $y_{lab}$ in Alg.1) are derived by performing a forced alignment procedure on the original training datasets. See the standard s5 recipe of Kaldi for more details \cite{kaldi}.

The weights of the network are initialized according to the \textit{Glorot} initialization \cite{xavier}, while biases are initialized to zero.
Training is based on a standard Stochastic Gradient Descend (SGD) optimization with mini-batches of size 128. The performance on the development set is monitored after each epoch and the learning rate is halved when the performance improvement is below a certain threshold. The training ends when no significant improvements have been observed for more than four consecutive epochs. 
The main hyperparameters of the system (i.e., learning rate, number of hidden layers, hidden neurons per layer, dropout factor and $\lambda$) have been optimized on the development set. 

The proposed system, which has been implemented with Theano \cite{theano}, 
has been coupled with the Kaldi toolkit \cite{kaldi} to form a context-dependent DNN-HMM speech recognizer.

\subsection{Relation to prior work}
Similarly to this paper, a joint training framework has been explored in \cite{joint2,joint1,joint3,joint6,joint7,joint4,joint5}. A key difference with previous works is that we propose to combine joint training with batch normalization. 
In \cite{joint1,joint3}, for instance, the joint training was actually performed as a fine-tuning procedure, which was carried out only after training the two networks independently. A critical aspect of such an approach is that the learning rate adopted in the fine-tuning step has to be properly selected in order to really take advantage of pre-training. With batch normalization we are able not only to significantly improve the performance of the system, but also to perform joint training from scratch, skipping any pre-training phase. 

Another interesting aspect of this work is a deeper study of the role played by the gradient weighting factor $\lambda$.


\section{Corpora and tasks}
\label{sec:corpora}
In order to provide an accurate evaluation of the proposed technique, the experimental validation has been conducted using different training datasets, different tasks and various environmental conditions\footnote{To allow reproducibility of the results reported in this paper,  the code of our joint-training system will be available at \url{https://github.com/mravanelli}. In the same repository, all the scripts needed for the data contamination will be available. The public distribution of the DIRHA-English dataset is under discussion with the Linguistic Data Consortium (LDC).}. 

The experiments with TIMIT are based on a phoneme recognition task (aligned with the Kaldi s5 recipe). The original training dataset has been contaminated with a set of realistic impulse responses measured in a real apartment. The reverberation time ($T_{60}$) of the considered room is about 0.7 seconds. Development and test data have been simulated with the same approach. More details about the data contamination approach can be found in \cite{IRs_paper,lrec,rav_is16}.

The WSJ experiments are based on the popular wsj5k task (aligned with the CHiME 3 \cite{chime3} task) and are conducted under two different acoustic conditions. For the \textit{WSJ-Rev} case, the training set is contaminated with the same set of impulse responses adopted for TIMIT. For the \textit{WSJ-Rev+Noise} case, we also added non-stationary noises recorded in a domestic context (the average SNR is about 10 dB). The test phase is carried out with the DIRHA English Dataset, consisting of 409 WSJ sentences uttered by six native American speakers in the above mentioned apartment. For more details see \cite{dirha_asru,rav_is16}.

\section{Experiments}
\subsection{Close-talking baselines}
\label{sec:ct_baseline}
The Phoneme Error Rate (PER\%) obtained by decoding the original test sentences of TIMIT is $19.5\%$ (using DNN models trained with the original dataset). The Word Error Rate (WER\%) obtained by decoding the close-talking WSJ sentences is $3.3\%$.  It is worth noting that, under such favorable acoustic conditions, the DNN model leads to a very accurate sentence transcription, especially when coupled with a language model.

\subsection{Joint training performance}
\label{sec:jt_pers}
In Table \ref{tab:res1}, the proposed joint training approach is compared with other competitive strategies. 
\begin{table}[t!]
\centering
\tabcolsep=0.28cm
    \begin{tabular}{ | l | c | c | c | c | }
    \cline{1-4}
    \multirow{2}{*}{\backslashbox{\em{System}}{\em{Dataset}}} & \multicolumn{1}{ | c |}{TIMIT}  & \multicolumn{1}{ | c |}{WSJ} & \multicolumn{1}{ | c |}{WSJ}  \\ \cline{2-4}
    & \textit{Rev} & \textit{Rev} & \textit{Rev+Noise}  \\ \hline
      Single big DNN & 31.5  & 8.1 & 14.3    \\ \hline
      SE + clean SR & 31.1  & 8.5 & 15.7    \\ \hline
      SE + matched SR & 30.1  & 8.0 & 13.7    \\ \hline
      SE + SR joint training & \textbf{29.2}  & \textbf{7.8} & \textbf{12.7}    \\ \hline  
    \end{tabular}
\caption{Performance of the proposed joint training approach compared with other competitive DNN-based systems.}
\label{tab:res1}
\end{table}
\label{sec:bn_exp}
\begin{table}[t!]
\centering
\tabcolsep=0.108cm
    \begin{tabular}{ | l | c | c | c | c | c |}
    \cline{1-5}
    \multirow{2}{*}{\backslashbox{\em{Dataset}}{\em{System}}} & \multicolumn{2}{ | c |}{Without Pre-Training}  & \multicolumn{2}{ | c |}{With Pre-Training}  \\ \cline{2-5}
    & \textit{no-BN} & \textit{with-BN} & \textit{no-BN} & \textit{with-BN}  \\ \hline
      TIMIT-Rev & 34.2  & 29.2 & 32.6  & 29.5   \\ \hline
      WSJ-Rev & 9.0  & 7.8 & 8.8  & 7.8   \\ \hline
      WSJ-Rev+Noise & 15.7 & 12.7 & 15.0  & 12.9  \\ \hline
    
    \end{tabular}
\caption{Analysis of the role played by batch normalization within the proposed joint training framework.}
\label{tab:test2}
\end{table}
In particular, the first line reports the results obtained with a single neural network. The size of the network has been optimized on the development set (4 hidden layers of 1024 neurons for TIMIT, 6 hidden layers of 2048 neurons for WSJ cases). The second line shows the performance obtained when the speech enhancement neural network (4 hidden layers of 2048 neurons for TIMIT, 6 hidden layers of 2048 neurons for WSJ) is trained independently and later coupled with the close-talking DNN of Sec.~\ref{sec:ct_baseline}. These results are particularly critical because, especially in adverse acoustic conditions, the speech enhancement model introduces significant distortions that a close-talking DNN trained in the usual ways is not able to cope with. To partially recover such a critical mismatch, one approach is to first train the speech enhancement, then pass all the training features though the speech enhancement DNN, and, lastly, train the speech recognition DNN with the dataset processed by the speech enhancement. The third line shows results obtained with such a matched training approach. The last line reports the performance achieved with the proposed joint training approach. Batch normalization is adopted for all the systems considered in Table \ref{tab:res1}.

Although joint training exhibits in all the cases the best performance, it is clear that such a technique is particularly helpful especially when challenging acoustic conditions are met. For instance, a relative improvement of about $8\%$ over the most competitive matched training system is obtained for the WSJ task in noisy and reverberant conditions.

\subsection{Role of batch normalization}
In Table \ref{tab:test2}, the impact of batch normalization on the joint training framework is shown.
The first two columns report, respectively, the results obtained with and without batch normalization when no pre-training techniques are employed. The impact of pre-training is studied in the last two columns. The pre-training strategy considered here consists of initializing the two DNNs with the matched training system discussed in Sec.~\ref{sec:jt_pers}, and performing a fine-tuning phase with a reduced learning rate. The column corresponding to the pre-training without batch normalization represents a system that most closely matches the approaches followed in \cite{joint1,joint3}. 

\begin{figure}
\centering
\includegraphics[width=0.52\textwidth]{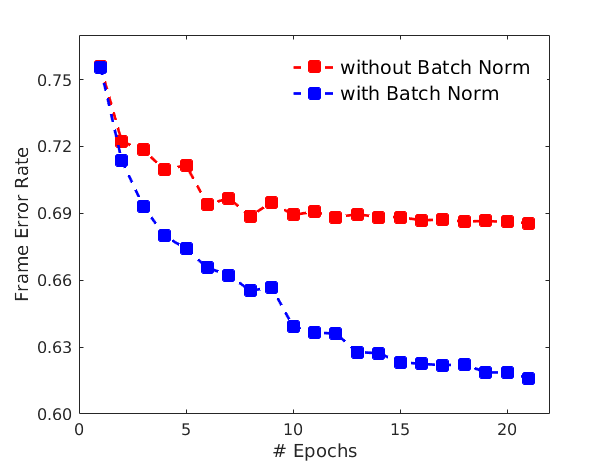}
\caption{Evolution of the test frame error rate across various training epochs with and without batch normalization.}
\label{fig:bn_frame}
\end{figure}

Table~\ref{tab:test2} clearly shows that batch normalization is particularly helpful. For instance, a relative improvement of about 23\% is achieved when batch normalization is adopted for the WSJ task in a noisy and reverberant scenario. The key importance of batch normalization is also highlighted in Fig.~\ref{fig:bn_frame}, where the evolution during training of the frame-level phone error rate (for the TIMIT-Rev dataset) is reported with and without batch normalization. From the figure it is clear that batch normalization, when applied to the considered deep joint architecture, ensures a faster convergence and a significantly better performance. Moreover, as shown in Table~\ref{tab:test2}, batch normalization eliminates the need of DNN pre-training, since similar (or even slightly worse results) are obtained when pre-training and batch normalization are used simultaneously.



\subsection{Role of the gradient weighting}
\label{sec:gw}
\begin{figure}
\centering
\includegraphics[width=0.49\textwidth]{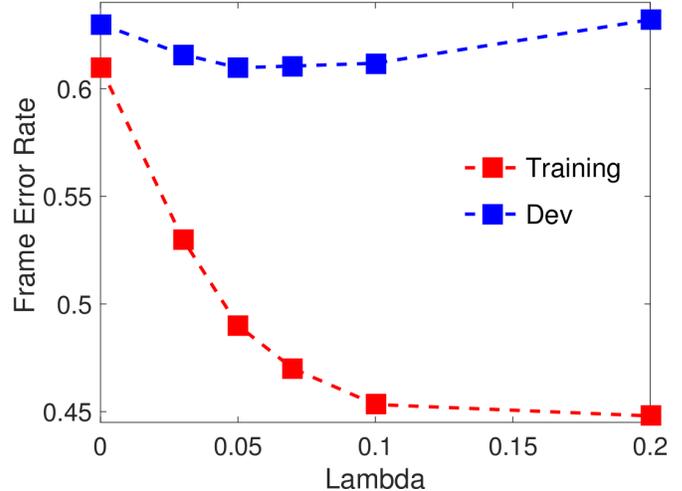}
\caption{Training and development frame error rates obtained on the TIMIT-Rev dataset for different values of $\lambda$.}
\label{fig:grad_w}
\end{figure}

In Fig. \ref{fig:grad_w}, the role of the gradient weighting factor $\lambda $ is highlighted.
From the figure one can observe that small values of $\lambda$ lead to a situation close to underfitting, while higher values of $\lambda$ cause overfitting. The latter result is somewhat expected since, intuitively, with very large values of $\lambda$ the speech enhancement information tends to be neglected and training relies on the speech recognition gradient only.

In the present work, we have seen that values of $\lambda$ ranging from 0.03 to 0.1 provide the best performance. Note that these values are smaller than that considered in \cite{joint1,joint2}, where a pure gradient summation ($\lambda=1$) was adopted. We argue that this result is due to the fact that, as observed in \cite{initbn}, the norm of the gradient decays very slowly when adopting batch normalization with a proper initialization of $\gamma$, even after the gradient has passed through many hidden layers. This causes the gradient backpropagated through the speech recognition network and into the speech enhancement network to be very large.



\section{Conclusion}
In this paper, a novel approach for joint training coupled with batch normalization is proposed. The experimental validation, conducted considering different tasks, datasets and acoustic conditions, showed that batch-normalized joint training is particularly effective in challenging acoustic environments, characterized by both noise and reverberation. In particular, batch normalization was of crucial importance for improving the system performance. A remarkable result is the relative improvement of about 23\% obtained for the WSJ task in a noisy and reverberant scenario when batch normalization is used within the joint training framework.

This system can be seen as a first step towards a better and more fruitful integration of the various technologies involved in current distant speech recognition systems. Future efforts for improving the current solution will be devoted to progressively involve different NN architectures or to embed other technologies such as speech separation, speaker identification and acoustic scene analysis.



\bibliographystyle{IEEEbib}
\bibliography{strings,refs}

\end{document}